\documentclass[conference,a4paper]{APSIPA2019}
\usepackage[psamsfonts]{amssymb}
\usepackage{amsxtra}
\usepackage{threeparttable}
\usepackage{amsfonts}
\usepackage{amssymb}
\usepackage{stfloats}
\usepackage{cite}
\usepackage{graphicx}
\usepackage{epstopdf}
\usepackage{psfrag}
\usepackage{subfigure}
\usepackage{amsmath}
\usepackage{array}
\usepackage{stfloats}
\usepackage{multirow}
\usepackage{color}
\usepackage{booktabs}
\usepackage{url}

\begin{document}

\title{A Study on Angular Based Embedding Learning for Text-independent Speaker Verification}

\author{%
\authorblockN{%
Zhiyong Chen, Zongze Ren and Shugong Xu
}
\authorblockA{%
Shanghai Institute for Advanced Communication and Data Science, Shanghai University, Shanghai, China\\
Email: \{bicbrv, zongzeren, shugong\}@shu.edu.cn
}
}

\maketitle
\thispagestyle{empty}

\begin{abstract}
Learning a good speaker embedding is important for many automatic speaker recognition tasks, including verification, identification and diarization. The embeddings learned by softmax are not discriminative enough for open-set verification tasks. Angular based embedding learning target can achieve such discriminativeness by optimizing angular distance and adding margin penalty. We apply several different popular angular margin embedding learning strategies in this work and explicitly compare their performance on Voxceleb speaker recognition dataset. Observing the fact that encouraging inter-class separability is important when applying angular based embedding learning, we propose an exclusive inter-class regularization as a complement for angular based loss. We verify the effectiveness of these methods for learning a discriminative embedding space on ASV task with several experiments. These methods together, we manage to achieve an impressive result with 16.5\% improvement on equal error rate (EER) and 18.2\% improvement on minimum detection cost function comparing with baseline softmax systems.
\end{abstract}

\section{Introduction}
Automatic speaker verification (ASV) is the task of determining the ID of a speaker given his/her voice. Lots of works have been focused on this open-set recognition task. In the past several years, the mainstream is to generatively model the speaker information base on manually designed acoustic features such as MFCC\cite{reynolds2000speaker, kenny2008study}. I-vector based system enables us to estimate a low-level utterance level feature given an acoustic feature of a sentence, using factor analysis\cite{dehak2010front}. Probabilistic linear discriminant analysis (PLDA)\cite{prince2007probabilistic, garcia2011analysis} models the speakers in the utterance level embedding space, enabling us to scoring the trials with a likelihood ratio.

Recently, the trend in this field is to use deep neural networks as the utterance level embedding extractor, exploiting the non-linearity and powerful representation power of DNN. This trend has lead to a wide range of studies that successfully improve the performance on ASV, such as the well-known x-vector/PLDA system\cite{snyder2017deep, snyder2018x}. Moreover, many studies are now focusing on exploring an end-to-end speaker verification system\cite{jung2019rawnet, jung2018complete, jung2018avoiding, wan2018generalized}, which use a neural network to model all component in a speaker verification system. 

Learning a better embedding is important for all recent DNN based ASV systems no matter what front-end or back-end we use. The object of embedding learning is to find a discriminative feature representation at a specific hidden layer of a neuron network. This requires to learn an embedding space that has good intra-class compactness and inter-class separability. Softmax cross entropy loss has been widely used in recognition and classification task including ASV\cite{variani2014deep, snyder2017deep, snyder2018x}. Triplet loss is designed to encourage the embeddings in the same class to have smaller Euclidean distance than embeddings from different classes\cite{schroff2015facenet}, which also have successfully implemented in ASV tasks\cite{zhang2017end, li2017deep}. However, there also remains some disadvantages to these methods. For softmax, the learned embeddings may be suitable for closed-set classification problems, but not discriminative enough for open-set verification problems. For triplet loss, triplet sampling method turns out to be tricky for effective model training.

In the face recognition field, the recently proposed large margin angular based embedding learning method is effective\cite{liu2017sphereface,wang2018cosface,deng2019arcface,wang2018additive}. These methods encourage the embeddings to compactly distributed in the unit hypersphere by adding an angular margin between embeddings and their cluster centers. Some of these methods have been exploited for the ASV task\cite{novoselov2018triplet,huang2018angular,liu2019large,li2018angular} and better results have been observed. 

In some more recent works, inter-class regularization is found to be effective when we are using angular based loss\cite{zhao2019regularface,liu2018learning,liu2017deep,duan2019uniformface}. These studies show that angular based softmax tend to focus mainly on intra-class compactness but not inter-class separability. This makes the embedding centers from different classes not well distributed in the embedding space when the dimension of embeddings are high. 

In this work, we give a more detailed study on the angular based embedding learning methods for text-independent ASV task comparing to existing works. We consider our contribution as follows: 1. Apply several different types of large margin angular based loss for ASV task and make an explicit comparative experiment for their performance. 2. Propose an angular based inter-class regularization which leads to improved results consistently, and verify the effect of such method with qualitative and quantitative experiments. We also share our findings of some effective training methods to give a successful implementation of angular based methods on ASV. These methods together lead to impressive improvements comparing with the baseline system.

\section{Angular Based Embedding Learning}
\subsection{Angular based softmax}
The most widely used classification loss function is known as softmax cross entropy, presented as follows:
\begin{eqnarray}
L_{softmax} = -\frac{1}{N}\sum_{i=1}^{N}\log{\frac{e^{\mathbf{w}_{y_i}^{T}\mathbf{x}_i+b_{y_i}}}{\sum_{j=1}^{C}e^{\mathbf{w}_{j}^{T}\mathbf{x}_i+b_j}}},
\label{softmax} 
\end{eqnarray}
where $ \mathbf{x}_i \in \Re^{d} $ denotes the last hidden layer output for the i-th sample, which is also known as an embedding. $y_i$ is the label of $\mathbf{x}_i$. The embedding feature dimension $d$ is set to 512 in this paper. $\mathbf{w}_j$ denotes the j-th column of the output linear matrix ${\mathbf{W} \in \Re^{d \times C}}$, and ${b_j \in \Re^C}$ is the bias term. $N$ and $C$ are batch size and class number. Following\cite{deng2019arcface,liu2017sphereface} we fix $\mathbf{b}_j = 0$ and transform the logit as $ \mathbf{w}_j^{T}\mathbf{x}_i = \lVert \mathbf{w}_j \rVert \lVert \mathbf{x}_i \rVert cos (\theta_{j, i}) $. We further normalize $\lVert \mathbf{w}_j \rVert = 1$ to have modified softmax loss:
\begin{eqnarray}
L_{modified} = -\frac{1}{N}\sum_{i=1}\log\frac{e^{\lVert \mathbf{x}_i \rVert cos(\theta_{y_i, i})}}{\sum_{j=1}e^{\lVert \mathbf{x}_i \rVert cos(\theta_{j, i})}},
\label{mod} 
\end{eqnarray}
in which $ \theta_{j, i} (0 \leqslant \theta_{j,i} \leqslant \pi)$ is the angle between vector $\mathbf{w}_j$ and $\mathbf{x}_i$. Note $\mathbf{w}_j$ can thus be seen as the \textit{cluster centers}. Modified softmax is able to directly optimize angles by making the decision boundary only depends on the angles. We do not further normalize the length of $\mathbf{x}_i$ and give a fix scale constraint in our study, since that is parameter sensitive, as suggested in \cite{liu2017deep}.

The recent studies show adding an angular margin to softmax is effective to learn more discriminative embeddings, and the connection with hypersphere manifold makes the learned features particularly suitable for open-set verification task. These methods can be concluded in the following equation:
\begin{equation}
\begin{split}
&L_{a} = \\
&-\frac{1}{N}\sum_{i=1}\log\frac{e^{\lVert \mathbf{x}_i \rVert (cos(m_1\theta_{y_i, i} + m_2) - m_3)}}{e^{\lVert \mathbf{x}_i \rVert (cos(m_1\theta_{y_i, i} + m_2) - m_3)}+\sum_{j \neq y_i}e^{\lVert \mathbf{x}_i \rVert cos(\theta_{j, i})}},
\label{angular}
\end{split}
\end{equation}
in which $m_1$, $m_2$ and $m_3$ are hyper-parameters to set angular margins. More specifically, setting $m_1$ gives angular softmax (A-softmax or SphereFace)\cite{liu2017sphereface}, $m_2$ gives additive angular margin softmax (AAM-softmax or ArcFace)\cite{deng2019arcface} and $m_3$ gives additive margin softmax (AM-softmax)\cite{wang2018additive}.

\subsection{Annealing}
Different from the implementation of angular based loss in face recognition, we observe some training difficulties when re-implementing these methods to ASV task. We find annealing is important for the convergence of all angular based loss in our study. For A-softmax system we use the following annealing as in \cite{liu2017sphereface}:
\begin{equation}
\begin{split}
f_{y_i} = \frac{\lambda \lVert \mathbf{x}_i \rVert cos(\theta_{y_i, i}) + \lVert \mathbf{x}_i \rVert cos(m\theta_{y_i, i})}{1 + \lambda},
\label{a_annealing}
\end{split}
\end{equation}
where $f_{y_i}$ is the $y_i$-th output logit given embedding $\mathbf{x}_i$, and $\lambda$ need to be gradually reduced during training. 

For AM-softmax and AAM-softmax systems, we find annealing is equivalently important to achieve neuron network convergence at the beginning of the training, we use a more straight forward method in these two cases:
\begin{equation}
\begin{split}
L_{a}^{'} = (1-\lambda^{'})L_{modified}+ \lambda^{'} L_{a},
\label{aam_annealing}
\end{split}
\end{equation}
where we gradually increase $\lambda^{'}$ in first several epochs to gradually shift the loss from modified softmax to large margin angular based loss.

\subsection{Inter class regularization}
Inter-class separability and intra-class compactness are two key factors to learn a discriminative embedding space. However, as suggested in \cite{zhao2019regularface, liu2018learning, duan2019uniformface}, current angular based embedding learning methods effectively encourage better intra-class compactness but the inter-class separability can be degraded due to feature redundancy. This may cause the \textit{cluster centers} not so well distributed in the high-dimension embedding space. Following the effectiveness in face recognition, we thus propose an angular regularization that explicitly encourages the \textit{cluster centers} in $\mathbf{W}$ to be uniformly distributed around the hypersphere. We consider the following separability measurement:
\begin{equation}
\begin{split}
SEP_{\mathbf{W}} = \frac{1}{C}\sum_{j}\sum_{i, i \neq j}max[0, cos(\phi_{i, j})]^{2},
\label{SEP}
\end{split}
\end{equation}
where $\phi_{i, j}$ is the angle between $\mathbf{w}_i$ and $\mathbf{w}_j$, ideally this scalar should be minimize to zero. 
This is equivalent to the following criterion:
\begin{equation}
\begin{split}
L_{inter} = \frac{1}{C} \lVert [\mathbf{W}_n^T\mathbf{W}_n]_{+} - \mathbf{I} \rVert_{F}^{2},
\label{F}
\end{split}
\end{equation}
where l2-normalized columns of $\mathbf{W}$ consist $\mathbf{W}_n$, $[\cdot]_{+}$ denotes clamping the matrix elements below zero, and squared Frobenius norm can be considered as calculating the energy of the matrix. We thus denote this angular regularization as \textit{hyperspherical energy} following\cite{liu2018learning}. We use the following loss function to combine the inter-class regularization:
\begin{equation}
\begin{split}
L_{a+inter} = (1-\lambda_{inter})L_{a}+ \lambda_{inter} L_{inter},
\label{a+inter}
\end{split}
\end{equation}
where $\lambda_{inter}$ is the hyper-parameter.

\section{Experiments}
\subsection{Experimental Setup}
Our experiment is based on the Voxceleb I \& II dataset\cite{nagrani2017voxceleb}\cite{chung2018voxceleb2}. To enhance the efficiency of experiments, we randomly sample 100 examples for each of the 5994 speakers in Voxceleb II development set to make a small training set for our experiment. Note we did not use any activity detection and data augmentation techniques to our training set following the baseline implementation\cite{nagrani2017voxceleb}. For testing, we use Voxceleb I verification trial list, consisting of 37720 non-target or target trials from 40 speakers. For each example, we use 512-point STFT as input feature. Mean and variance normalization on each frequency bin is performed.

The backbone deep neural network is similar to the baseline  Resnet structure used in \cite{chung2018voxceleb2}, but we use a more efficient Resnet18 structure to speed up training. Preactivate Resnet blocks\cite{he2016identity} are used in our implementation.

The major testing criterion is to evaluate equal error rate and minimum detection cost function (minDCF) with target prior set to 0.01. Since our focus is on the discriminative power from embedding learning, we use cosine distance scoring as a simple back-end. The length variability is dealt with averaging pooling on the hidden layer.

\subsection{Training Details}
We train our neural network with SGD optimizer, the initial learning rate is set to 0.1 in all of our experiments and step decay to 0.001, annealing techniques are used in all the angular based training. Note besides annealing, we find the choice of the large learning rate and sample balance is also critical to a successful implementation of angular based systems on ASV. We set $\lambda_{inter}$ to 0.01 and our implementation is based on Pytorch.

\begin{table*}[t]
\begin{center}
\begin{threeparttable}
\caption{Results of systems on Voxceleb}
\label{Results}
\begin{tabular}{|p{50mm}|p{20mm}|p{20mm}|p{20mm}|}
\hline
\rule[-5pt]{0pt}{16pt}Systems & Margin & EER & MinDCF(0.01)\\
\hline
Resnet34 Softmax\cite{chung2018voxceleb2} & - & 5.04 & 0.543 \\
\hline
Resnet18 Softmax (Baseline) & - & 5.33 & 0.489 \\
\hline
Resnet18 Modified Softmax & $m_1=1$ & 5.74 & 0.515 \\
\hline
\hline
Resnet18 AM-Softmax & $m_3=0.1$ & 5.17 & 0.485 \\
\hline
Resnet18 AM-Softmax & $m_3=0.2$ & 4.56 & 0.402 \\
\hline
Resnet18 AM-Softmax & $m_3=0.3$ & 4.81 & 0.456 \\
\hline
Resnet18 AM-Softmax + inter & $m_3=0.2$ & \underline{{\bfseries4.45}} & \underline{{\bfseries0.400}} \\ 
\hline
\hline
Resnet18 AAM-Softmax & $m_2=0.2$ & 4.63 & 0.446 \\
\hline
Resnet18 AAM-Softmax & $m_2=0.3$ & 4.55 & 0.443 \\
\hline
Resnet18 AAM-Softmax & $m_2=0.4$ & 4.99 & 0.485 \\
\hline
Resnet18 AAM-Softmax + inter & $m_2=0.3$ & \underline{4.49} & \underline{0.441} \\ 
\hline
\hline
Resnet18 A-Softmax & $m_1=2$ & 4.58 & 0.429 \\ 
\hline
Resnet18 A-Softmax & $m_1=3$ & 4.64 & 0.431 \\
\hline
Resnet18 A-Softmax & $m_1=4$ & 4.75 & 0.468 \\
\hline
Resnet18 A-Softmax + inter & $m_1=2$ & \underline{4.46} & \underline{0.427} \\ 
\hline
\end{tabular}
\end{threeparttable}
\end{center}
\end{table*}


\subsection{Results and Analysis}
In Table~\ref{Results}, the performance of the systems using softmax and different kinds of angular based loss functions are compared. We use the corresponding type-1 results given in\cite{chung2018voxceleb2} as the baseline. From our tested results, the Resnet18 and our small training set are sufficient to achieve reasonable results and conduct the experiments. The modified softmax system perform worse than softmax, due to the pruned bias term. AM-softmax, AAM-softmax and A-softmax systems with different margins are explicitly compared. The angular based systems perform consistently better than softmax, showing the effectiveness of the adding large angular margin. Comparing different angular based systems, we see that their improvements to the baseline are similar. Note that adding relatively small margin is enough to get ideal improvement on our ASV task. We get better performance if all these angular based losses are regularized with inter-class separability. We manage to achieve the best performance with AM-softmax with inter-class regularization at margin 0.2 among all the system we implement.

\begin{figure}[!b] \centering 
\includegraphics[height=7cm,width=8cm]{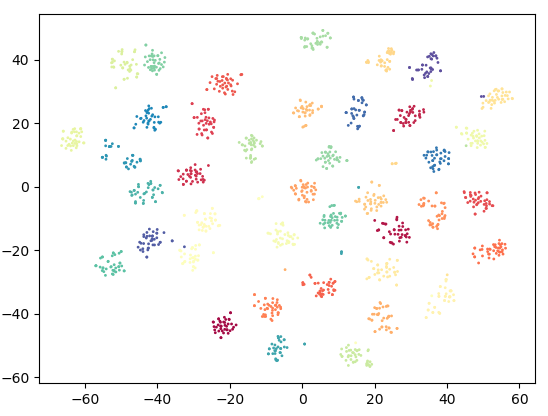}  
\includegraphics[height=7cm,width=8.1cm]{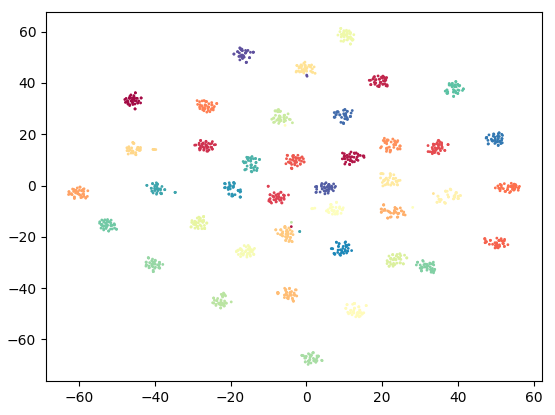}  
\caption{Embedding plot of 40 classes from Voxceleb training data. Top: Softmax, Bottom: Angular based softmax. The embeddings are dimension reduced using t-SNE.} 
\label{emb}
\end{figure}

To further explore the effectiveness of angular based systems, we plot the embeddings of 40 different classes draw from the training set, using t-SNE to reduce dimension Fig.~\ref{emb}. We see that by training the neural network we obtain a linear separable embedding space. Within each class, the samples distribute more compactly when we use AM-softmax (m=0.2) comparing with the softmax, which clearly shows that angular based loss encourage better intra-class compactness.

\begin{table}[t]
\begin{center}
\begin{threeparttable}
\caption{Comparing on inter-class separability}
\label{sep}
\begin{tabular}{|p{40mm}|p{15mm}|p{15mm}|}
\hline
\multicolumn{1}{|c|}{\rule[-5pt]{0pt}{16pt}System} & $SEP_{\mathbf{W}}$ & $S_b$ \\
\hline
Resnet18 Softmax & 25.8 & 0.910 \\ 
\hline
Resnet18 A-Softmax & 39.3 & 0.907 \\ 
\hline
Resnet18 A-Softmax + inter & {\bfseries22.8} & {\bfseries0.942} \\ 
\hline
\end{tabular}
\end{threeparttable}
\end{center}
\end{table}

\begin{figure}[b] \centering 
\includegraphics[height=6cm,width=8.5cm]{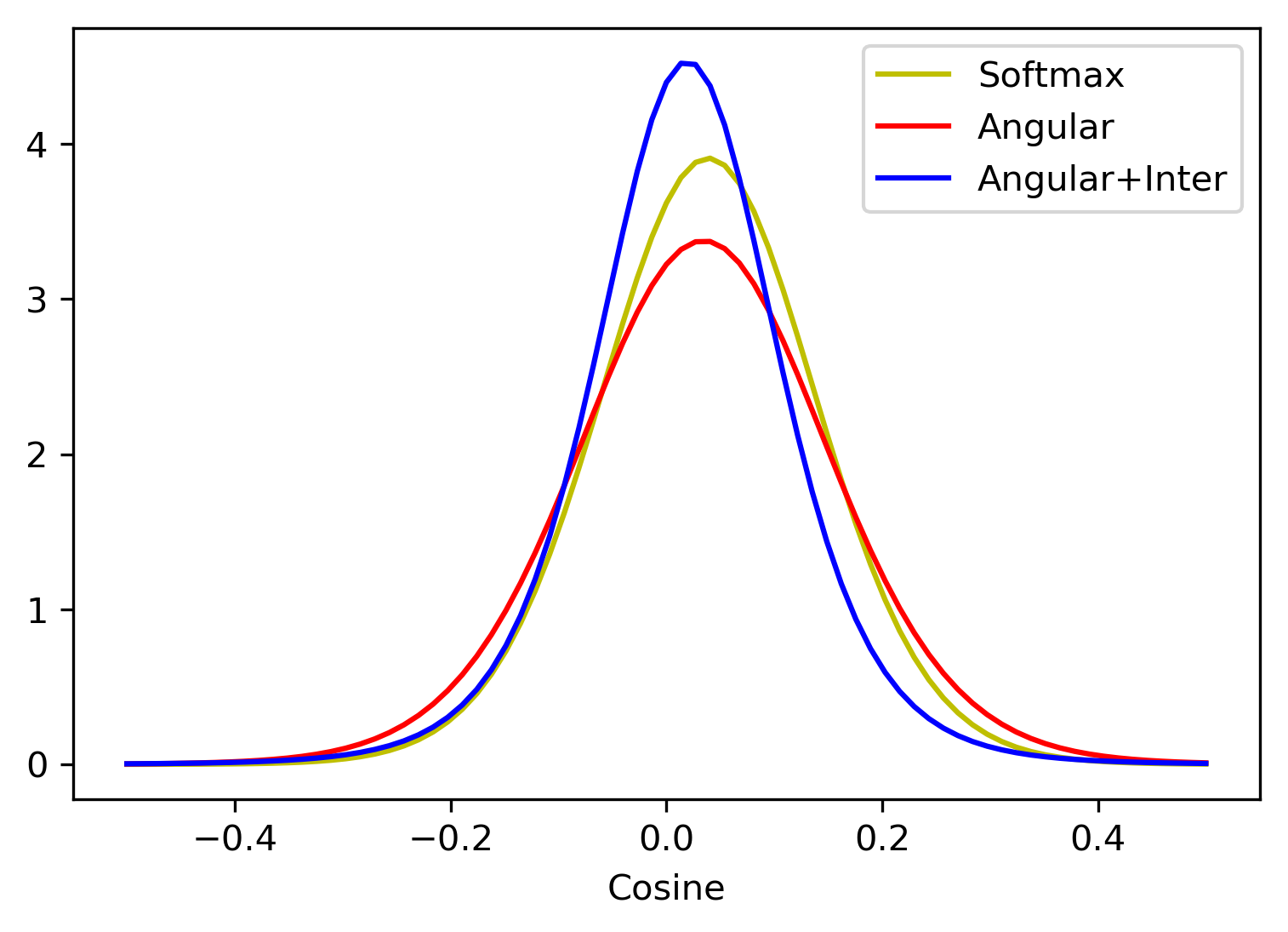} 
\caption{Score distribution plot of the Voxceleb non-target trials, fit by Student's t-distributions. Inter-class regularization makes the scores compactly distributed near zero.} 
\label{score}
\end{figure}

We also explore the effectiveness of adding the inter-class regularization. In Talbe~\ref{sep}, we first compare the \textit{hyperspherical energy} in equation~\ref{SEP}, which reflect the separability between the \textit{cluster centers}. We see that using A-softmax, the \textit{cluster centers} become less separably distributed in the embedding space, even comparing with the softmax system, which is not ideal to fully utilize the embedding space to learn better class separable embeddings. This motivate us to explicitly combining the inter-class regularization as a complement with A-softmax, which lead to reduced \textit{hyperspherical energy}. This shows such regularization encourages inter-class separability. We also evaluate the between-class angular variance $S_b$ like\cite{liu2017sphereface} given by,
 \begin{eqnarray}
S_b = \frac{1}{N}\frac{1}{C-1}\sum_{i}^{C}n_i\sum_{j, j \neq i}^{C}(1-cos \langle m_i, m_j \rangle),
\label{Sb} 
\end{eqnarray}
on the testing set. N is the sample number, C is the class number of the testing set, $m_{i}$ is the mean vector from class i. A-softmax with inter-class regularization has larger between-class angular variance comparing to other systems. As shown in Fig.~\ref{score}, we plot the score distribution from the Voxceleb non-target trials, which is a good indicator for the distance between examples from different classes in the testing set. We calculate their cosine distances and use a Student's t-distribution to fit results of each system. The A-softmax system has large score variance and many embeddings from different classes stay too close on the hypersphere. With inter-class regularization, the scores of non-target trails tend to distribute closer to zero with a smaller variance, indicating a large proportion of the angels between embeddings from different classes are 90 degrees. These experiments show that inter-class regularizing not only lead to a better-separated \textit{cluster centers} but also well generalize to the class separability on the testing set. 

\begin{figure}[t] \centering 
\includegraphics[height=7cm,width=8.5cm]{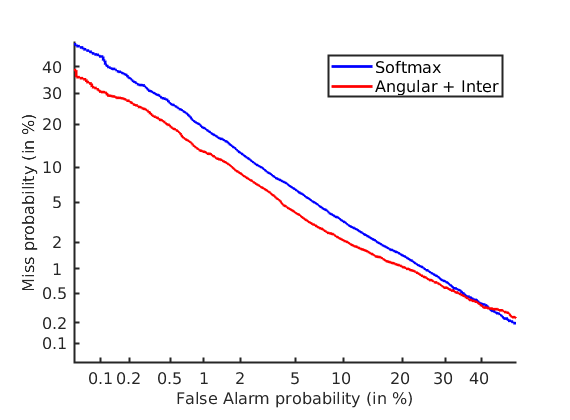} 
\caption{DET plot for softmax system and angular based softmax system.} 
\label{det}
\end{figure}

We finally draw the detection error trade-off (DET) plot of the softmax system and inter-class regularized AM-softmax system in Fig.~\ref{det}. We see that the proposed system outperform the baseline softmax system in almost all operation points. The gap between the two systems is especially large at lower false alarm range, suggesting that the angular based system can have superior performance on open-set speaker verification tasks. 

\section{Conclusions \& Future Works}
We have explored a sort of angular based embedding learning methods on ASV task. By adding a large angular margin, we manage to achieve better intra-class compactness. And with inter-class regularization, we effectively learned a more class-separable space. All these methods lead to more discriminative speaker embeddings and better speaker verification performance. The performance of several angular based learning methods has been compared on Voxceleb dataset. Angular based loss with inter-class regularization achieves apparently better results comparing with baseline softmax. The effectiveness of the inter-class regularization has also been studied, which improves the inter-class separability of training cluster centers and generalizes to testing data. In the future, we plan to conduct more detailed experiments on different strategies to add inter-class regularization. We will also conduct experiments on more datasets to see the robustness of these methods.

\clearpage
\bibliographystyle{IEEEtran}
\bibliography{mybib}

\end{document}